\documentclass[conference]{IEEEtran}

\IEEEoverridecommandlockouts
\usepackage{cite}
\usepackage{amsmath,amssymb,amsfonts}
\usepackage{mathtools}
\usepackage{algorithmic}
\usepackage{graphicx}
\usepackage{textcomp}
\usepackage{amsfonts}
\usepackage{xcolor}
\usepackage{times}
\usepackage{fancyhdr}
\usepackage[ruled,vlined]{algorithm2e}

\def\doubleunderline#1{\underline{\underline{#1}}}

\def\BibTeX{{\rm B\kern-.05em{\sc i\kern-.025em b}\kern-.08em
    T\kern-.1667em\lower.7ex\hbox{E}\kern-.125emX}}

\DeclareMathOperator{\new}{new}
\DeclareMathOperator{\old}{old}
\DeclareMathOperator{\intn}{int}
\DeclareMathOperator{\ext}{ext}
\DeclarePairedDelimiterX{\infdivx}[2]{(}{)}{%
  #1\;\delimsize\|\;#2%
}

\DeclarePairedDelimiter{\norm}{\lVert}{\rVert}

\begin{document}

\title{Unsupervised Fuzzy eIX: \\ Evolving Internal-eXternal Fuzzy Clustering}

\author{\IEEEauthorblockN{Charles Aguiar}
\IEEEauthorblockA{\textit{Department of Automatics} \\
\textit{Federal University of Lavras (UFLA)}\\
Minas Gerais, Brazil \\
charlescaguiar@hotmail.com}
\and
\IEEEauthorblockN{Daniel Leite}
\IEEEauthorblockA{\textit{Department of Automatics} \\
\textit{Federal University of Lavras (UFLA)}\\
Minas Gerais, Brazil \\
daniel.leite@ufla.br}
}

\maketitle

\begin{abstract}
Time-varying classifiers, namely, evolving classifiers, play an important role in a scenario in which information is available as a never-ending online data stream. We present a new unsupervised learning method for numerical data called evolving Internal-eXternal Fuzzy clustering method (Fuzzy eIX). We develop the notion of double-boundary fuzzy granules and elaborate on its implications. Type 1 and type 2 fuzzy inference systems can be obtained from the projection of Fuzzy eIX granules. We perform the principle of the balanced information granularity within Fuzzy eIX classifiers to achieve a higher level of model understandability. Internal and external granules are updated from a numerical data stream at the same time that the global granular structure of the classifier is autonomously evolved. A synthetic nonstationary problem called Rotation of Twin Gaussians shows the behavior of the classifier. The Fuzzy eIX classifier could keep up with its accuracy in a scenario in which offline-trained classifiers would clearly have their accuracy drastically dropped.

\end{abstract}


\section{Introduction}

Real-world data streams usually undergo changes over time, which portray the evolution of the environment they come from. To conceive, represent and handle information, a computational model should be able to adapt itself in response to changes of the underlying process or phenomenon. The ability of self-adjustment consists in updating the model parameters and structure to track unknown events, behavioral patterns, and gradual and abrupt changes of the system operating conditions. Inference and learning algorithms able to notice changes in input data streams and update model parameters and structure to keep a synopsis of the current data profile have been largely studied \cite{vskrjanc2019evolving, angelov2013evolving, OverView, leite2012evolving}.

Fuzzy models can be constructed and updated by means of online incremental algorithms. As discussed in \cite{vskrjanc2019evolving, angelov2013evolving, OverView, leite2012evolving, bouchachia2007overview,Cordovil}, evolving models are characterized by:

\begin{itemize}
    \item ability to learn online, i.e., the model is updated, generally in an instance-per-instance basis, as data are available;
    \item ability to gradually tune its parameters and structure to track concept drifts and shifts; and
    \item needlessness of \textit{a priori} knowledge about the data, data properties, and amount of classes/patterns.
\end{itemize}

\noindent Online learning should trade off structural plasticity and stability. In other words, learning procedures should parsimoniously reconcile and decide on either creating new granules or updating existing ones. Structural plasticity means creating new granules to memorize new concepts. Plasticity avoids learned granules to be exposed to catastrophic forgetting. Structural stability preserves the model structure, but allows adaptation of existing granules to smooth and gradual changes. Usually, online machine learning and data mining algorithms are not effective in finding such equilibrium because methods to assess the current relevance of information granules, balance the size of granules along the problem dimensions, and merge similar granules are not considered.

Storage of large volumes of data from nonstationary environments is very often infeasible or ineffective \cite{Souza, SilvaP, pratama2018evolving}. Evolving intelligent systems provide an autonomous approach for data stream analysis. These systems usually comprise a single-scan-through-the-data learning method, which is of utmost importance in time-critical, high-frequency, nonstationary, and big data applications. Furthermore, evolving systems offer an open platform in which local components, e.g., granules, neurons, clusters, leaves or rules, can be automatically generated, updated, merged, split, and recalled based on the behaviour of the data stream \cite{tung2013et2fis, Lana, angelov2013evolving}. Evolving models have shown to be \textit{self organizing} since no previous knowledge about the data is needed \cite{Cordovil,SilvaA,vskrjanc2019evolving,lughofer2011evolving}.

In addition to change over time, data streams produced from electronic devices and real-world perceptions consist of inherently uncertain values. Uncertainty is a feature that indicates how much a measured value deviates from the true value. Uncertainty can originate from fluctuations of random nature in data flows, numerical imprecision due to binary computation, fusion of information from different sources, non-ideal measuring instruments, data pre-processing methods, and others \cite {leite2012evolving,Lio,Garcia,liu2007uncertainty}. Evolving granular models particularly capable of learning from interval and fuzzy data streams, and incorporate and take advantage of data uncertainties are given in \cite{EGNN,FBEM,OWA}. Most evolving models, however, learn from numerical data streams while the representation of the data is given by fuzzy or interval objects (local models) \cite{vskrjanc2019evolving}.

This paper presents a new online learning framework called evolving Internal-eXternal Fuzzy clustering method, or Fuzzy eIX for short. A Fuzzy eIX model is useful for classification. Different from any other evolving approach, a Fuzzy eIX model is formed by double-boundary information granules extracted from an unsupervised numerical data stream. The granules contain an internal structure which summarizes the statistics of a fraction of the dataset. Therefore, local internal information is used for autonomous decision-making along the online learning process. The double-boundary feature of Fuzzy eIX granules can be used to translate clustering results into a type-1 or type-2 fuzzy inference system at any time. Additionally, Fuzzy eIX performs the principle for a balanced information granularity \cite{bargiela2016granular}, which affirms that granules should be balanced along all dimensions for a better general understandability of the results in an application domain. Therefore, Fuzzy eIX is strongly aligned with the concept of eXplainable Artificial Intelligence (XAI), which says that the solutions provided by machine learning algorithms and models should be understood by humans.

The remainder of this paper is structured as follows. Section II reviews some related online unsupervised algorithms. Section III describes practical implications of developing double-boundary information granules. Section IV outlines the Fuzzy eIX framework for unsupervised numerical data-stream modeling. We show how internal and external boundaries arise and are updated from the data to represent uncertainties. Section V gives preliminary results on a synthetic nonstationary problem called Rotation of Twin Gaussians \cite{leite2012evolving,EGNN}. Section VI concludes the paper.

\section{Related Literature}

We address some related studies on evolving clustering and briefly discuss some studies that somehow go into the idea of double-boundary granulation.

\subsection{Evolving Clustering}

Historically, the evolving Clustering Method (eCM) \cite {song2001ecm} was the first evolving approach to the clustering problem. ECM is based on the Euclidean distance to define the similarity between new data and clusters. A distance threshold between an instance and the center of a cluster is used to compute membership levels. A new cluster is created if an instance is sufficiently distant to all clusters. Cluster centers are dragged toward instances that have a significant membership degree in the cluster. Although ECM updates clusters' centers and radii and creates new clusters on the fly, it does not delete, split and combine clusters.

SOStream is an evolving clustering algorithm based on self-organizing density maps \cite{isaksson2012sostream}. The algorithm processes a new instance similar to the ECM algorithm. However, at each iteration SOStream checks for clusters that overlap at a given $\gamma$ level, and merge them. Clusters are created until a minimum value of a meta-parameter is reached. The ability to merge and update clusters structure based on the data flow gives SOStream the property of self-organization. However, if little or nothing is known about the data, it is impractical to provide an assertive value for the minimum number of clusters to be generated. An excessive number of clusters may be created to represent the data. This issue is aggravated since the approach is not supplied with deleting mechanism.

Typicality and Eccentricity-based Data Analysis (TEDA) is a framework by \cite{angelov2014outside} that introduces the concept of \textit{eccentricity} and \textit{typicality}. Eccentricity means how distinct a particular data instance is from other instances and from the current focal points of clouds. Typicality is based on how similar a data instance is in relation to the entire data set and a cloud \cite{Bezerra}. These concepts provide measures of density and proximity. For each new instance, the eccentricity and typicality are recursively calculated to determine if an instance is either typical or anomalous. TEDA establishes data density levels that allow the identification of anomalies based on similarity levels. TEDA is applicable to fault and outlier detection problems, and prediction. In \cite{soares2018ensemble}, a variation of the TEDA method is given for weather prediction. 

\subsection{Double-Boundary Granulation}

Pawlak rough sets \cite{pawlak1982rough} can be used to perform approximate classification of uncertain and inaccurate data. A rough set is defined by lower and upper approximation sets, which can be crisp or fuzzy.  The region between the lower and upper sets is the boundary region -- a region in which a point may or may not belong to the set. A rough set is a uni-granular construct, i.e., the boundary of one knowledge granule is the issue for the definition of a rough model \cite{tripathy2015multigranular}. However, many studies have extended the original ideas toward multi-granular constructs. 

Rough-set-based models for information granulation are addressed in \cite{yao2001information}. Optimistic multi-granular rough sets are proposed in \cite{qian2006rough}. A pessimistic multi-granular rough set approach for problem solving in the context of multiple granulation is presented in \cite{qian2010pessimistic}. Several properties and extensions of optimistic and pessimistic  multi-granulation methods are described in \cite{tripathy2014multi, tripathy2014covering, nagaraju2015approximate}.

\section{Why Fuzzy eIX Clustering?}

We envision implications of developing Fuzzy eIX granules from online data streams.

\subsection{Type-1 Evolving Fuzzy Inference System}

\begin{figure}[htbp]
\centerline{\includegraphics[scale=0.65]{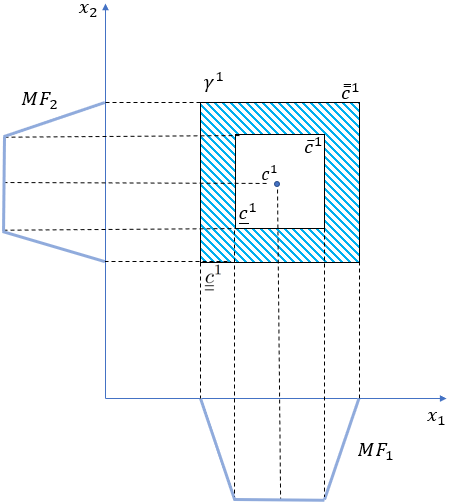}}
\caption{Evolving type-1 membership functions and a type-1 inference system from Fuzzy eIX double-boundary granules}
\label{fig3}
\end{figure}

By projecting the internal and external boundaries of Fuzzy eIX granules in orthogonal axes representing the attributes of a problem, trapezoidal membership functions and, therefore, an evolving type-1 fuzzy inference system, are obtained. As granules are created and updated in the Cartesian product space, the core and support of associated membership functions evolve from the data stream. Figure \ref{fig3} shows an example. Naturally, the overall Fuzzy eIX model can be read linguistically from a set of If-Then rules -- a rule per granule.

\subsection{Type-2 Evolving Fuzzy Inference System}

\begin{figure}[htbp]
\centerline{\includegraphics[scale=0.65]{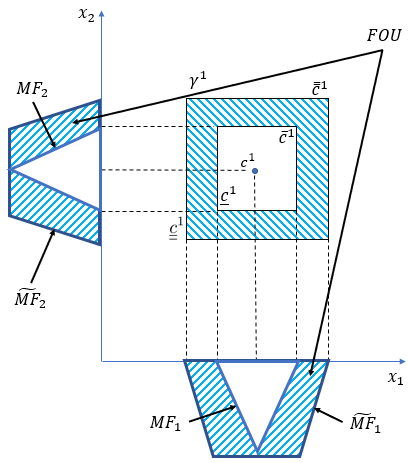}}
\caption{Evolving type-2 membership functions and a type-2 inference system from Fuzzy eIX double-boundary granules}
\label{fig4}
\end{figure}

A rigorous way to project a Fuzzy eIX granule in orthogonal axes consists in considering the midpoint of a granule as the prototype, and the information of the internal granule only. Rigorous membership functions that describe the uncertainty related to the proximity of a point to the prototypical point of a granule are established by means of the inner boundaries of the granule. An evolving type-2 fuzzy inference system can be obtained from the inner and outer projections, yielding $MF_i$ and $\widetilde{MF}_i$, respectively, as shown in Fig. \ref{fig4}. The hatched region between membership functions is called footprint of uncertainty (FOU). The larger the FOU area, the greater the uncertainty, and vice-versa. As the parameters of $MF_i$ and $\widetilde{MF}_i$ are straightly obtained from evolving granules, they autonomously learn values for themselves from the data stream.

\subsection{Local and Global Inter-Granular Uncertainty}

\begin{figure}[htbp]
\centerline{\includegraphics[scale=0.6]{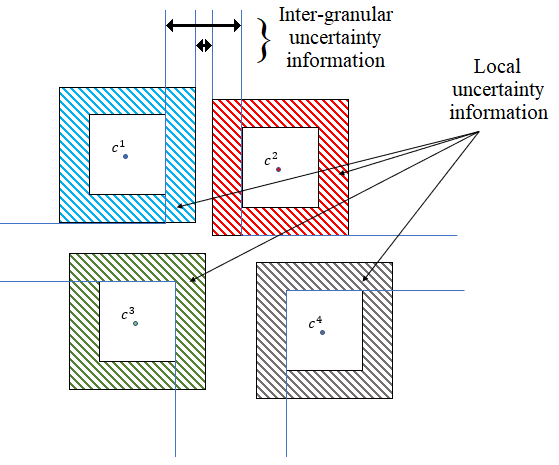}}
\caption{Inter-granular and local uncertainty information in Fuzzy eIX granules}
\label{fig6}
\end{figure}

A hyper-rectangular double-boundary granule contains four parameters per problem dimension, which are the bounds of internal and external intervals. Updating intervals according to proportions looking to the values of a single attribute only is a straightforward and fast way of learning in dynamic environment. In addition to local uncertainty representation given by the hatched regions of Fig. \ref{fig6}, information about inter-granular uncertainty can also be computed and used to make decisions within an evolving learning algorithm. For example, multiple merging and conflict-solving procedures can be derived by comparing  double-boundary granules.

\section{Fuzzy eIX: Learning Algorithm}

Fuzzy eIX granules are delimited by inner and outer hyper-boxes. The membership degree of an instance that belongs to the inner region of a granule is 1. An instance belonging to the area between the inner and outer boxes is partially considered a member of the granule. Central points and inner and outer bounds of granules are recursively updated over time according to a data stream.

\subsection{Initialization}

Let a data stream be denoted by $\textbf{x}^{[h]}$, $h = 1, ...$. Given the first instance, $\textbf{x}^{[1]} \in \mathbb {R}^{n}$, namely, $\textbf{x}^{[1]} = [x_1^{[1]} ~ ... ~ x_j^{[1]} ~ ... ~ x_n^{[1]}]$ -- being $n$ the number of attributes -- the first granule $\gamma^1$ is created. Its center, $\textbf{c}^1$, is equal to $\textbf{x}^{[1]}$. The initial widths of the inner and outer intervals of $\gamma^1$ are equal to $\epsilon$ and $2\epsilon$, respectively, in any dimension $j$, being $\epsilon \in [0, 0.5]$ a meta-parameter. Therefore, the inner and outer bounds are given by $[\underline{\textbf{c}}^1, \overline{\textbf{c}}^1]$ and $[\doubleunderline{\textbf{c}}^1, \overline{\overline{\textbf{c}}}^1]$, respectively. The vector of centers and inner and outer bounds are adaptive over time.

A generic granule $\gamma^i$ of a collection $\gamma = \{\gamma^1, ..., \gamma^k\}$ is characterized by:

\begin{itemize}
    \item a prototypical point or center, $\textbf{c}^i \in \mathbb{R}^{n}$; and
    \item lower and upper inner bounds, $\underline{\textbf{c}}^i$, $\overline{\textbf{c}}^i \in \mathbb{R}^{n}$, and lower and upper outer bounds, $\underline{\underline{\textbf{c}}}^i$, $\overline{\overline{\textbf{c}}}^i \in \mathbb{R}^{n}$.
\end{itemize}

\begin{figure}[htbp]
\centerline{\includegraphics[scale=0.64]{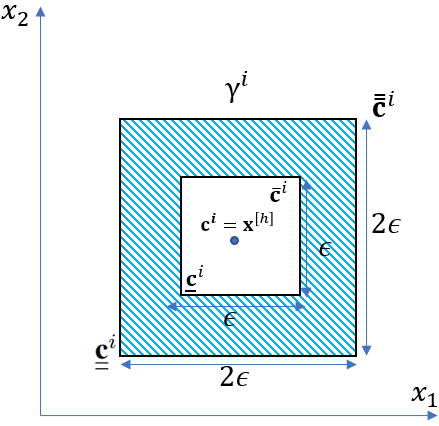}}
\caption{A generic bi-dimensional Fuzzy eIX granule, $\gamma^i$}
\label{fig1}
\end{figure}

\noindent Additionally, the meta-parameter $\epsilon$ defines the initial and minimum possible width for the attributes of granules. The meta-parameter $\rho$ is useful to the merging procedure, as described later. Thus, $\gamma^i \coloneqq \{\textbf{c}^i,\underline{\textbf{c}}^i, \overline{\textbf{c}}^i,\doubleunderline{\textbf{c}}^i, \overline{\overline{\textbf{c}}}^i\}$ is a five-fold collection of $n$-dimensional vectors. Figure \ref{fig1} shows a bi-dimensional example.

\subsection{Instance in Inner Region}

If an instance $\textbf{x}^{[h]}$ is placed within the limits of the inner region of a granule $\gamma^i$, i.e.,

\begin{equation}
    \label{test1}
  \begin{split}
    x^{[h]}_j > \underline{c}^i_j \\
    x^{[h]}_j < \overline{c}^i_j
  \end{split}
\quad
  \begin{split}
    , ~ \forall j, ~ j = 1,...,n,
  \end{split}
\end{equation}

\noindent then the membership degree of $\textbf{x}^{[h]}$ in $\gamma^i$ is 1 by means of any T-norm. As the certainty on the placement of $\gamma^i$ is greater with the inclusion of $\textbf{x}^{[h]}$, internal and external widths become smaller. The lower inner bound is increased from

\begin{align}
    \begin{split}
        \underline{c}^i_j(\new) &= (1 + d^{i[h]}_j) ~ \underline{c}^i_j(\old), \label{eq2}
    \end{split}
\end{align}

\noindent $j = 1, ..., n$, with $d^{i[h]}_j \in [0, \beta]$ given by

\begin{align}
    \begin{split}
        d^{i[h]}_j &:= \beta - \beta \left( \frac{|c^i_j - x_j^{[h]}|}{c^i_j - \underline{c}_j^i} \right). \label{eq3}
    \end{split}
\end{align}

\noindent The default value of $\beta$ is 0.3. The lower outer bound, $\underline{\underline{c}}^i_j$, is increased analogously to \eqref{eq2}-\eqref{eq3}. The upper inner bound, $\overline{c}^i_j$, is reduced proportionally from

\begin{align}
    \begin{split}
        \overline{c}^i_j(\new) &= \overline{c}^i_j(\old) - \left( \underline{c}^i_j(\new) - \underline{c}^i_j(\old) \right), \label{eq4}
    \end{split}
\end{align}

\noindent $j = 1, ..., n$. Similarly, the upper outer bound, $\overline{\overline{c}}^i_j$, is obtained from \eqref{eq4} using the lower outer bounds in the last term. Notice that the shrinkage is larger when $\textbf{x}^{[h]}$ is closer to $\textbf{c}^i$.

The size of Fuzzy eIX granules is limited such that the width of the internal and external regions, in any dimension, is greater than $\epsilon$ and $2\epsilon$, respectively, at any iteration. 

After shrinking the $i$-th granule, the granule slides toward the current instance inversely proportional to its density, i.e., to the number of instances, $N^i$, that belonged to the inner region of $\gamma^i$ previously (a weighted drift).

The center is moved using

\begin{equation}
    \textbf{c}^i(\new) = \textbf{c}^i(\old) + \frac{1}{N^i+1}(\textbf{x}^{[h]} - \textbf{c}^i(\old)).
\end{equation}

\noindent Let $\mathcal{\bullet}^i$ be any boundary $\doubleunderline{\textbf{c}}^i$, $\underline{\textbf{c}}^i$, $\overline{\textbf{c}}^i$,  $\overline{\overline{\textbf{c}}}^i$. To preserve symmetry of the hyper-rectangular shape, inner and outer boundaries move accordingly,

\begin{equation}
    \mathcal{\bullet}^i(\new) = \mathcal{\bullet}^i(\old) + \frac{1}{N^i+1}(\textbf{x}^{[h]} - \textbf{c}^i(\old)).
\end{equation}

\begin{figure}[h]
\centerline{\includegraphics[scale=0.5]{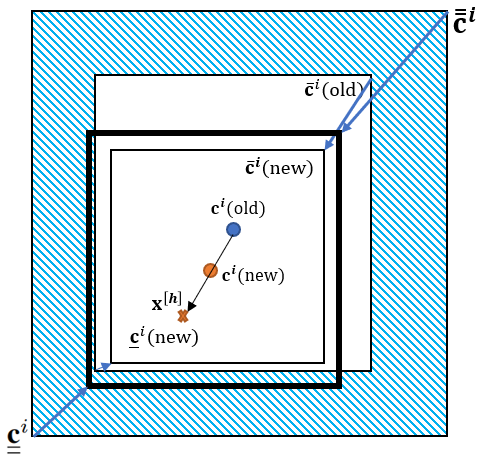}}
\caption{Sliding and shrinking the $i$-th granule as the current instance $x^{[h]}$ belongs to its inner region and therefore increase the certainty of its location}
\label{fig5a}
\end{figure}

Figure \ref{fig5a} shows an example of the shrinking and sliding procedures due to the current instance $x^{[h]}$, which belongs to its inner region of $\gamma^i$.

\color{black}

\subsection{Instance in Outer Region}

An instance $\textbf{x}^{[h]}$ may belong to the outer granule, with one or more of its attributes placed outside the inner region. In this case,

\begin{equation}
    \label{test2}
  \begin{split}
    x^{[h]}_j > \doubleunderline{c}^i_j \\
    x^{[h]}_j < \overline{\overline{c}}^i_j
  \end{split}
\quad
  \begin{split}
    , ~ \forall j, ~ j = 1,...,n.
  \end{split}
\end{equation}

\noindent In addition to the feasibility of \eqref{test2}, Eq. \eqref{test1} must be false for at least one dimension, $j$. The $L_\infty$ distance from $\textbf{x}^{[h]}$ to $\textbf{c}^i$ is greater than the distance between $\textbf{c}^i$ and any instance within the inner bounds of $\gamma^i$. 

Instances belonging to the outer region of $\gamma^i$ imply that the center $\textbf{c}^i$ should not be drifted as we are not sure about its content. The uncertainty on the membership of $\textbf{x}^{[h]}$ in $\gamma^i$ rather suggests granular expansion for inclusion. 

Internal and external widths become larger as follows. The lower inner bound is reduced from

\begin{align}
    \begin{split}
        \underline{c}^i_j(\new) &= (1 - f^{i[h]}_j) ~ \underline{c}^i_j(\old), \label{eq8}
    \end{split}
\end{align}

\noindent with

\begin{align}
    \begin{split}
        f^{i[h]}_j &:= \beta \left( \frac{\underline{c}^i_j - x_j^{[h]}}{\underline{c}^i_j - \underline{\underline{c}}_j^i} \right), \label{eq9}
    \end{split}
\end{align}

\noindent considering only the dimensions $j$ in which $\underline{\underline{c}}_j^i \leq x_j^{[h]} \leq \underline{c}_j^i$. Otherwise, $\underline{c}^i_j$ is not changed. The default value of $\beta$ is 0.3; and $f^{i[h]}_j \in [0, \beta]$. The lower outer bound, $\underline{\underline{c}}^i_j$, is reduced analogously, using \eqref{eq8} and the same $f^{i[h]}_j$ obtained in \eqref{eq9}. 

The upper inner bound, $\overline{c}^i_j$, is increased proportionally from

\begin{align}
    \begin{split}
        \overline{c}^i_j(\new) &= \overline{c}^i_j(\old) + \left( \underline{c}^i_j(\old) - \underline{c}^i_j(\new) \right) \label{eq10}
    \end{split}
\end{align}

\noindent $j = 1, ..., n$. Similarly, the upper outer bound, $\overline{\overline{c}}^i_j$, is obtained from \eqref{eq10} using the lower outer bounds in the last term.

Additionally, for the dimensions $j$ in which $\overline{c}_j^i \leq x_j^{[h]} \leq \overline{\overline{c}}_j^i$, Eq. \eqref{eq8} is applied using $\overline{c}_j^i$, instead of $\underline{c}_j^i$; and $f^{i[h]}_j$ is got from

\begin{align}
    \begin{split}
        f^{i[h]}_j &:= - \beta \left( \frac{x_j^{[h]} - \overline{c}^i_j}{\overline{\overline{c}}_j^i - \overline{c}^i_j} \right). \label{eq11}
    \end{split}
\end{align}

\noindent In this case, $f^{i[h]}_j \in [-\beta, 0]$. The upper outer bound, $\overline{\overline{c}}^i_j$, is increased by analogy, i.e., using \eqref{eq8}, and $f^{i[h]}_j$ as in \eqref{eq11}. Moreover, the lower inner bound, $\underline{c}^i_j$, is reduced proportionally from

\begin{align}
    \begin{split}
        \underline{c}^i_j(\new) &= \underline{c}^i_j(\old) - \left( \overline{c}^i_j(\new) - \overline{c}^i_j(\old) \right) \label{eq12}
    \end{split}
\end{align}

\noindent $j = 1, ..., n$. Similarly, the lower outer bound, $\underline{\underline{c}}^i_j$, is obtained from \eqref{eq12} using the upper outer bounds in the last term. Notice that the expansion is larger as the attributes of $\textbf{x}^{[h]}$ approach the outer borders.

\begin{figure}[htbp]
\centerline{\includegraphics[scale=0.5]{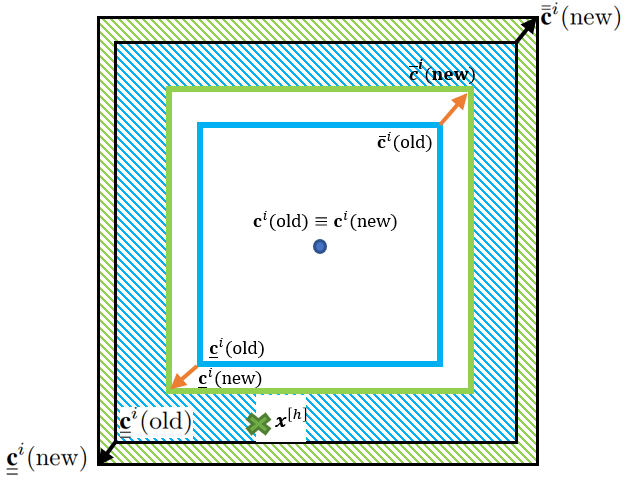}}
\caption{Granular expansion as a consequence of the instance $\textbf{x}^{[h]}$ be included in the outer region of $\gamma^i$, but not included in its inner region}
\label{fig5b}
\end{figure}

The size of Fuzzy eIX granules is limited such that the width of the internal and external regions, in any dimension and at any iteration, is greater than $\epsilon$ and $2\epsilon$, respectively. Figure \ref{fig5b} shows an example of granular expansion in reaction to the evidence $\textbf{x}^{[h]}$, which belongs to the outer region of $\gamma^i$.

\subsection{Eccentric Instance}

If $\textbf{x}^{[h]}$ does not belong to any internal and external regions of all granules, a new granule is created to include $\textbf{x}^{[h]}$. The granule creation procedure is similar to that when the first instance arises. The new granule grasps the new information, brought by $\textbf{x}^{[h]}$.

\subsection{Balanced Information Granularity}

The balanced granularity principle  \cite{bargiela2016granular} states that preference should be given to the development of local objects with balanced granularity along the different axes. This means that the width of the lines that delineate a hyper-rectangle should be ideally similar. A better understandability of the results in an application domain can be reached by means of a balanced granular model.

Fuzzy eIX internal and external granules are balanced at each iteration based on average widths. Let

\begin{align}
    \label{ind}
    \begin{split}
        w^{i(\intn)}_{j} &:= \overline{c}^i_j - \underline{c}^i_j,
    \end{split}
    \begin{split}
         ~w^{i(\ext)}_{j} &:= \overline{\overline{c}}^i_j - \underline{\underline{c}}^i_j,
    \end{split}
\end{align}

\noindent $j = 1, ..., n$, be the internal and external widths, respectively. 

Let $k$ be the current amount of granules. For individual attributes $j$, $j = 1, ..., n$, the internal and external average width of all granules along the $j$-th axis are 

\begin{align}
    \label{mean}
    \begin{split}
        \hat{w}^{(\intn)}_j &= \frac{1}{k}\sum_{i=1}^k ~ w^{i(\intn)}_{j},
    \end{split}
    \begin{split}
        ~\hat{w}^{(\ext)}_j &= \frac{1}{k}\sum_{i=1}^k ~ w^{i(\ext)}_{j}.
    \end{split}
\end{align}

\noindent Individual widths \eqref{ind} are updated toward average widths \eqref{mean} considering one attribute at a time. Relatively smaller sides of hyper-rectangles are gradually expanded whereas larger sides are reduced. Formally,

\begin{align}
    \begin{split}
        \doubleunderline{c}^i_j(\new) &= \doubleunderline{c}^i_j(\old) - \alpha (\hat{w}^{(\ext)}_j-w^{i(\ext)}_{j}) \\
        \underline{c}^i_j(\new) &= \underline{c}^i_j(\old) - \alpha (\hat{w}^{(\intn)}_j-w^{i(\intn)}_{j}) \\
        \overline{c}^i_j(\new) &= \overline{c}^i_j(\old) + \alpha (\hat{w}^{(\intn)}_j-w^{i(\intn)}_{j}) \\
        \overline{\overline{{c}}}^i_j(\new) &= \overline{\overline{{c}}}^i_j(\old) + \alpha (\hat{w}^{(\ext)}_j-w^{i(\ext)}_{j})
    \end{split}
\end{align}

\noindent $\forall i,j$; $i = 1, ..., k$; $j = 1, ..., n$, in which $\alpha \in [0,1]$ is the balancing rate. We set $\alpha = 0.3$ as default value. Higher values of $\alpha$ increase the speed of convergence of granules to a similar size, and provide higher model interpretability at the price of a potential lost in accuracy related to the existence of classes with different spreads. If accuracy is the most important aspect, then, smaller values of $\alpha$ keep the original hyper-rectangular geometry of granules and different spreads.

\subsection{Weighted Mean and Convex Hull Merging}

Merging happens when a pair of granules is notably overlapped. Often, a sequence of data instances belongs to the gap between granules, which used to be disjoint at a former time instant. Therefore,  redundancy is avoided by merging them \cite{vskrjanc2019evolving}. When two granules, say $\gamma^{i_1}$ and $\gamma^{i_2}$, are close enough, i.e., when their centers, $\textbf{c}^{i_1}$ and $\textbf{c}^{i_2}$, are such that

\begin{equation}
    \label{merge}
    \norm{\textbf{c}^{i_1} - \textbf{c}^{i_2}}_\infty \leq \rho,
\end{equation}

\noindent with $\rho \in [0,1]$, then they are merged. 

We introduce two merging methods. In the Weighted Mean method, the center of the new granule, $\gamma^{k+1}$, is

\begin{equation}
    \textbf{c}^{k+1} = \textbf{c}^{i_1} - \frac{N^{i_2}}{N^{i_1}+N^{i_2}}\left(\textbf{c}^{i_1} - \textbf{c}^{i_2}\right), \label{cm}
\end{equation}

\noindent in which $N^i$ is the number of times the $i$-th granule was chosen to be updated for a given input instances. The new granule is placed on the line between $\textbf{c}^1$ and $\textbf{c}^2$, and depends on the data density previous granules used to represent. The internal, $\underline{\textbf{c}}^{k+1}$, $\overline{\textbf{c}}^{k+1}
$, and external, $\doubleunderline{\textbf{c}}^{k+1}$, $\overline{\overline{\textbf{c}}}^{k+1}$, endpoints of the new granule, $\gamma^{k+1}$, are obtained from \eqref{cm} by analogy. 

\begin{figure}[b]
\centerline{\includegraphics[scale=0.4]{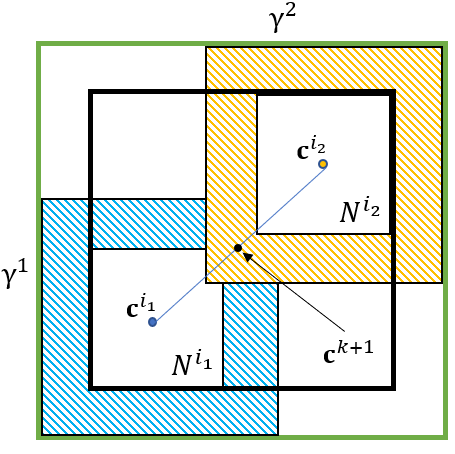}}
\caption{Convex hull approach to merge Fuzzy eIX granules}
\label{fig2}
\end{figure}

An alternative merging approach, the Convex Hull method, consists in producing a coarser granule that encapsulates all information inherent to the previous granules. In this case, the center of the new granule arises naturally from operations on endpoints. Namely, let

\begin{align}
    \begin{split}
        \doubleunderline{c}^{k+1}_j &= \min \{\doubleunderline{c}^{i_1}_j, \doubleunderline{c}^{i_2}_j \} \\
        \underline{c}^{k+1}_j &= \min \{\underline{c}^{i_1}_j, \underline{c}^{i_2}_j \} \\
        \overline{c}^{k+1}_j &= \max \{\overline{c}^{i_1}_j, \overline{c}^{i_2}_j \} \\
        \overline{\overline{c}}^{k+1}_j &= \max \{ \overline{\overline{c}}^{i_1}_j, \overline{\overline{c}}^{i_2}_j \}
    \end{split}
\end{align}

\noindent $j = 1, ..., n$. Then,

\begin{equation}
\textbf{c}^{k+1} = \frac{1}{2} \left(\overline{\textbf{c}}^{k+1} - \underline{\textbf{c}}^{k+1}\right)
\end{equation}

\noindent is the midpoint of the granule. At first sight, this merging approach increases the coverage area of the model, and preserves past information. The granule tends to be shrunk afterwards, toward average widths. Figure \ref{fig2} exemplifies the convex-hull merging procedure.

\subsection{Summary}

The Fuzzy eIX unsupervised algorithm is summarized below. The resulting classifier is parametrically and structurally adaptive. Hence, it deals with nonstationary data streams.

\begin{algorithm}[htbp]
\SetAlgoLined
\KwResult{$\{\gamma\}$: a set of granules}
$\gamma \leftarrow [~]$, $c = 0$\\
$\alpha = \beta = 0.3$\\
\ForEach{$\textbf{x}^{[h]}$, $h = 1, ...$ (data stream)}{
    \uIf{$h=1$}{
        $\gamma^{c+1} \leftarrow$ MakeGranule($\textbf{x}^{[h]}$, $\epsilon$)
     }
    \Else{
        \ForEach{$\gamma^i \in \gamma$}{
            \uIf{$\textbf{x}^{[h]} \in \gamma^i.Internal$}{
                ShrinkInternal($\textbf{x}^{[h]}$, $\gamma^i$, $\epsilon$)\\
                ShrinkExternal($\textbf{x}^{[h]}$, $\gamma^i$, $\epsilon$)\\
                SlideCenter($\textbf{x}^{[h]}$, $\gamma^i$)
            }
            \uElseIf{$\textbf{x}^{[h]} \in \gamma^i.External$}{
                ExpandInternal($\textbf{x}^{[h]}$, $\gamma^i$)\\
                ExpandExternal($\textbf{x}^{[h]}$, $\gamma^i$)
            }
            \Else{
                $\gamma^{c+1} \leftarrow$ MakeGranule($\textbf{x}^{[h]}$, $\epsilon$)
            }
        }
    }
    MergeClusters($\gamma$, $\rho$)\\
    BalanceGranules($\gamma$, $\alpha$)
  }
 \caption{FuzzyEIX ($\epsilon$, $\rho$)}
\end{algorithm}

\section{Preliminary Results}

The experiment called Rotation of the Twin Gaussians \cite{leite2012evolving} is a nonstationary classification problem useful to evaluate the effectiveness of the Fuzzy eIX method. 

A data stream is generated from two partially-overlapped Gaussian functions. The Gaussians are initially centered at $g_1^{[0]} = (4,4)$ and $g_2^{[0]} = (6,6)$, and have standard deviation of $0.8$. They rotate around the point $(5,5)$ according to

\begin{align}
    \begin{split}
        \theta_i^{[h]} &= \theta_i^{[h-1]} + \phi \\
        g_{i(1)}^{[h]} &= 5 + 1\sqrt{2}\cos(\theta_i^{[h]}) \\
        g_{i(2)}^{[h]} &= 5 + 1\sqrt{2}\sin(\theta_i^{[h]}),
    \end{split}
\end{align}

\noindent in which $\theta_i$ is the counterclockwise angle, around $(5,5)$, of the position of the center of the $i$-th Gaussian. Initially, as the centers are $(4,4)$ and $(6,6)$, then $\theta_1^{[0]} = 45$ degrees for the `Class 1', and $\theta_2^{[0]} = 225$ degrees for the `Class 2'. We consider two stages. First, the rotating rate, $\phi$, is $0$ during $h = 200$ time steps (stationary Gaussians). Then, the rotating rate, $\phi$, is $0.45$ from $h = 201$ to $h = 400$ (concept drift). An instance is produced from one of the Gaussians per time step. 

The data is scaled in [0,1]. We assume that the first class is the positive class. Consider a confusion matrix consisting of two rows and two columns, which represent the number of true positives (TP), false positives (FP), true negatives (TN), and false negatives (FN) \cite{fawcett2006introduction}. The accuracy of a classifier is obtained from

\begin{equation}
    Acc(\%) = \frac{TP + TN}{TP + FP + TN + FN} ~.~ 100\%.
\end{equation}

Table \ref{tab1} shows the Fuzzy eIX results for the stationary ($h =$ $1, ..., 200$) and nonstationary ($h = 201, ..., 400$) stages, considering the convex hull merging procedure and different initial meta-parameters.

\begin{table}[htbp]
\caption{Fuzzy eIX - Rotation of the Twin Gaussians}
\begin{center}
\begin{tabular}{cccc}
\hline
& \hspace{-70pt} \textbf{Stationary stage (200 instances)} \\
\hline
Parameters $\{\epsilon, \rho\}$ & $Acc$ (\%) & Avg. Granules & Time (s) \\
\hline
$\{0.035, 0.25\}$ &  85.0 & 6.01 & 0.26 \\
\hline
$\{0.045, 0.35\}$ &  88.0 & 5.78 & 0.10 \\
\hline
$\{0.055, 0.45\}$ & 94.5 & 3.12 & 0.18 \\
\hline
$\{0.065, 0.55\}$ & 84.5 & 4.41 & 0.21 \\
\hline
$\{0.055, 0.55\}$ & 90.0 & 3.86 & 0.19 \\
\hline
$\{0.065, 0.55\}$ & 81.0 & 3.07 & 0.24 \\
\hline
& \hspace{-70pt} \textbf{Nonstationary stage (200 instances)} \\
\hline
Parameters $\{\epsilon, \rho\}$ & $Acc$ (\%) & Avg. Granules & Time (s) \\
\hline
$\{0.035, 0.25\}$ & 81.0 & 12.78 & 0.45 \\
\hline
$\{0.045, 0.35\}$ & 83.0 & 9.32 & 0.41 \\
\hline
$\{0.055, 0.45\}$ & 90.5 & 5.69 & 0.36 \\
\hline
$\{0.065, 0.55\}$ & 79.5 & 4.12 & 0.31 \\
\hline
$\{0.055, 0.45\}$ & 85.5 & 5.43 & 0.35 \\
\hline
$\{0.065, 0.55\}$ & 77.5 & 4.25 & 0.29 \\
\hline
\end{tabular}
\label{tab1}
\end{center}
\end{table}

\begin{figure}[h]
	\centering
		\begin{tabular}{cc}
		\includegraphics[scale=0.36]{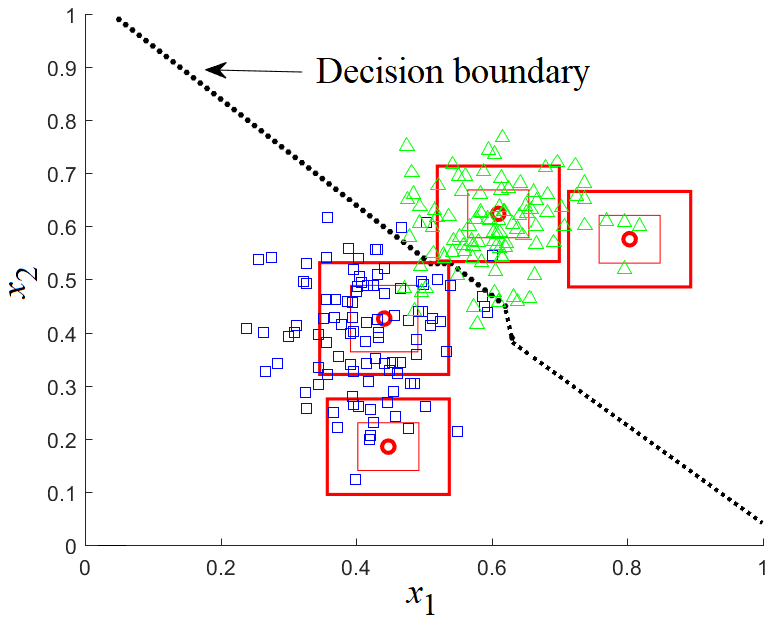} \\
		(a) Stationary stage \\ 
		\includegraphics[scale=0.58]{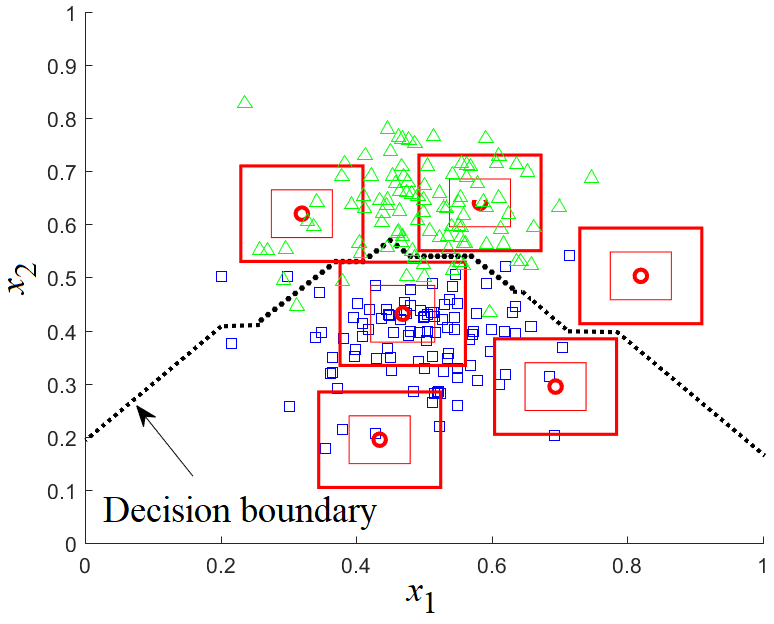} \\
		(b) Nonstationary stage \\
		\end{tabular}
	\caption{\label{results} Final position of Fuzzy eIX granules for the best setting of meta-parameters, $\epsilon = 0.055$, $\rho = 0.45$}
\end{figure}

\noindent For the parameters $\epsilon = 0.055$ and $\rho = 0.45$, an accuracy of $94.5\%$ and $90.5\%$ using $3.12$ and $5.69$ granules, for the stationary and nonstationary stages, respectively, were reached. For higher values of $\epsilon$, the accuracy is degraded as larger granules are assigned to instances belonging to different classes in the initial time steps. An average of $2.5$ additional granules is needed to keep the accuracy up during the gradual rotation of the data sources. Figure \ref{results} shows the Fuzzy eIX local elements, and the decision boundary for the best classifier.

\begin{figure}[h]
\centerline{\includegraphics[scale=0.3]{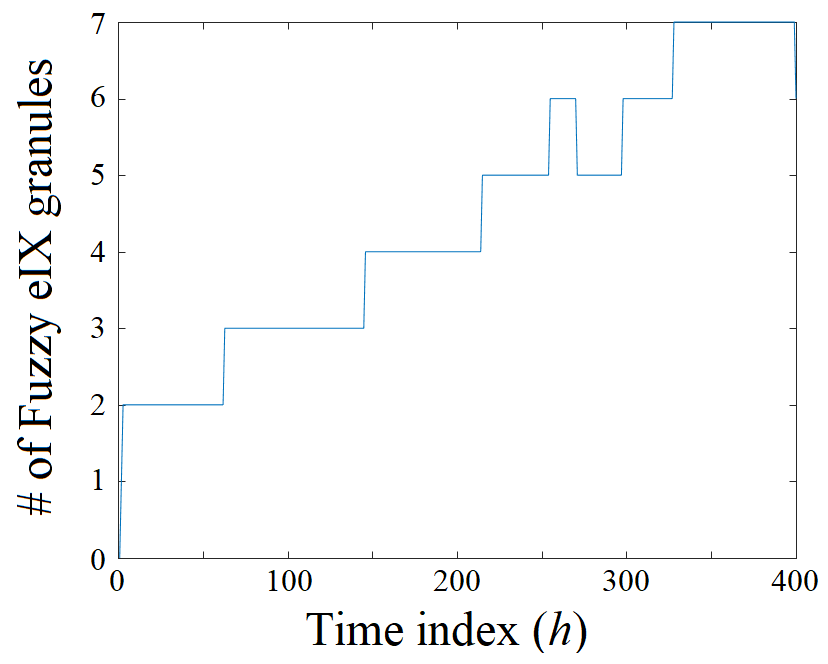}}
\caption{Evolution of the structure of the Fuzzy eIX classifier}
\label{granules}
\end{figure}

Figure \ref{granules} shows the evolution of the number of granules over time for the best classifier. Notice that the amount of granules increases generally faster after $h=200$ to address the classes drift. Granules are created on the fly to cover new instances. From Table \ref{tab1}, it is also noteworthy that the model's accuracy is not significantly affected by the concept change since structural and parametric changes are carried out by the Fuzzy eIX algorithm.

\section{Conclusion and Perspectives}

We proposed an unsupervised evolving granular framework that uses double-boundary granules and numerical data streams to construct classification models. The framework is called Fuzzy eIX. Local internal and external regions are updated recursively for instances with partial or full membership in a granule. Data properties and model structure are unknown beforehand, which gives a Fuzzy eIX classifier the characteristic of self-organization. Type 1 and type 2 fuzzy inference systems can be obtained from the projection of Fuzzy eIX granules in orthogonal axes. Moreover, the rule-based systems obtained is linguistically understandable in an application since the Fuzzy eIX learning algorithm balances the granules along the problem dimensions. Encouraging results have been obtained, such as results in the time-varying Rotation of the Twin Gaussians problem. The classifier could keep its accuracy in a certain level (90.5 - 94.5\%) in a scenario in which offline trained classifiers would clearly have their accuracy drastically dropped.

Several possible extensions of the Fuzzy eIX framework can be mentioned:

\begin{itemize}
    \item Granules' initial dimensions as well as minimum acceptable dimensions can be time-varying by updating the key meta-parameter, $\epsilon$;
    \item The weighted sliding procedure takes place if an instance lays within the inner bounds of a granule. Other relations should be considered to set weights, e.g., local width and uncertainty information;
    \item Merging methods should be further analysed and compared considering hyper-volumes, weighted means, and optimistic and pessimistic similarities. For instance, Hausdorff distance should be considered, as in \cite{jeng2019ipfcm};
    \item Heuristics will be evaluated to update the meta-parameter $\alpha$ to achieve faster or slower balancing speed;
    \item Results of Fuzzy eIX will be compared with those of other evolving classifiers using benchmark datasets;
    \item Interval and fuzzy data streams, and weak supervision shall be addressed in the future.
\end{itemize}

\bibliographystyle{ieeetr}

\bibliography{references}

\end{document}